\documentclass[10pt,twocolumn,letterpaper]{article}

\usepackage{iccv}
\usepackage{times}
\usepackage{epsfig}
\usepackage{graphicx}
\usepackage{amsmath}
\usepackage{amssymb}
\usepackage{multirow}

\usepackage[breaklinks=true,bookmarks=false]{hyperref}
\iccvfinalcopy
\def\iccvPaperID{****}

\ificcvfinal\fi

\begin{document}

\title{Global-Local Temporal Representations For Video Person Re-Identification}

\author{Jianing Li\textsuperscript{1}, Jingdong Wang\textsuperscript{1}, Qi Tian\textsuperscript{2}, Wen Gao\textsuperscript{1}, Shiliang Zhang\textsuperscript{1}\\
\normalsize{\textsuperscript{1}School of Electronics Engineering and Computer Science, Peking University}\\
\normalsize{\textsuperscript{2}Huawei Noahs Ark Lab}
}

\maketitle
\ificcvfinal\thispagestyle{empty}\fi
\begin{abstract}
This paper proposes the Global-Local Temporal Representation (GLTR) to exploit the multi-scale temporal cues in video sequences for video person Re-Identification (ReID). GLTR is constructed by first modeling the short-term temporal cues among adjacent frames, then capturing the long-term relations among inconsecutive frames. Specifically, the short-term temporal cues are modeled by parallel dilated convolutions with different temporal dilation rates to represent the motion and appearance of pedestrian. The long-term relations are captured by a temporal self-attention model to alleviate the occlusions and noises in video sequences. The short and long-term temporal cues are aggregated as the final GLTR by a simple single-stream CNN. GLTR shows substantial superiority to existing features learned with body part cues or metric learning on four widely-used video ReID datasets. For instance, it achieves Rank-1 Accuracy of 87.02\% on MARS dataset without re-ranking, better than current state-of-the art. 
\end{abstract}

\section{Introduction} \label{sec:introduction}
Person Re-Identification aims to identify a probe person in a camera network by matching his/her images or video sequences and has many real applications, including smart surveillance and criminal investigation.
Image person ReID has achieved significant progresses in terms of both solutions~\cite{su2017pose,li2019pose,li2018harmonious} and large benchmark dataset construction~\cite{li2014deepreid,zheng2015scalable,wei2018person}.
Recently, video person ReID, the interest of this paper, has been attracting a lot of attention because the availability of video data is easier than before, and video data provides richer information than image data. Being able to explore plenty of spatial and temporal cues, video person ReID has potentials to address some challenges in image person ReID, \eg, distinguishing different persons wearing visually similar clothes.

\begin{figure}
\centering
\includegraphics[width=1\linewidth]{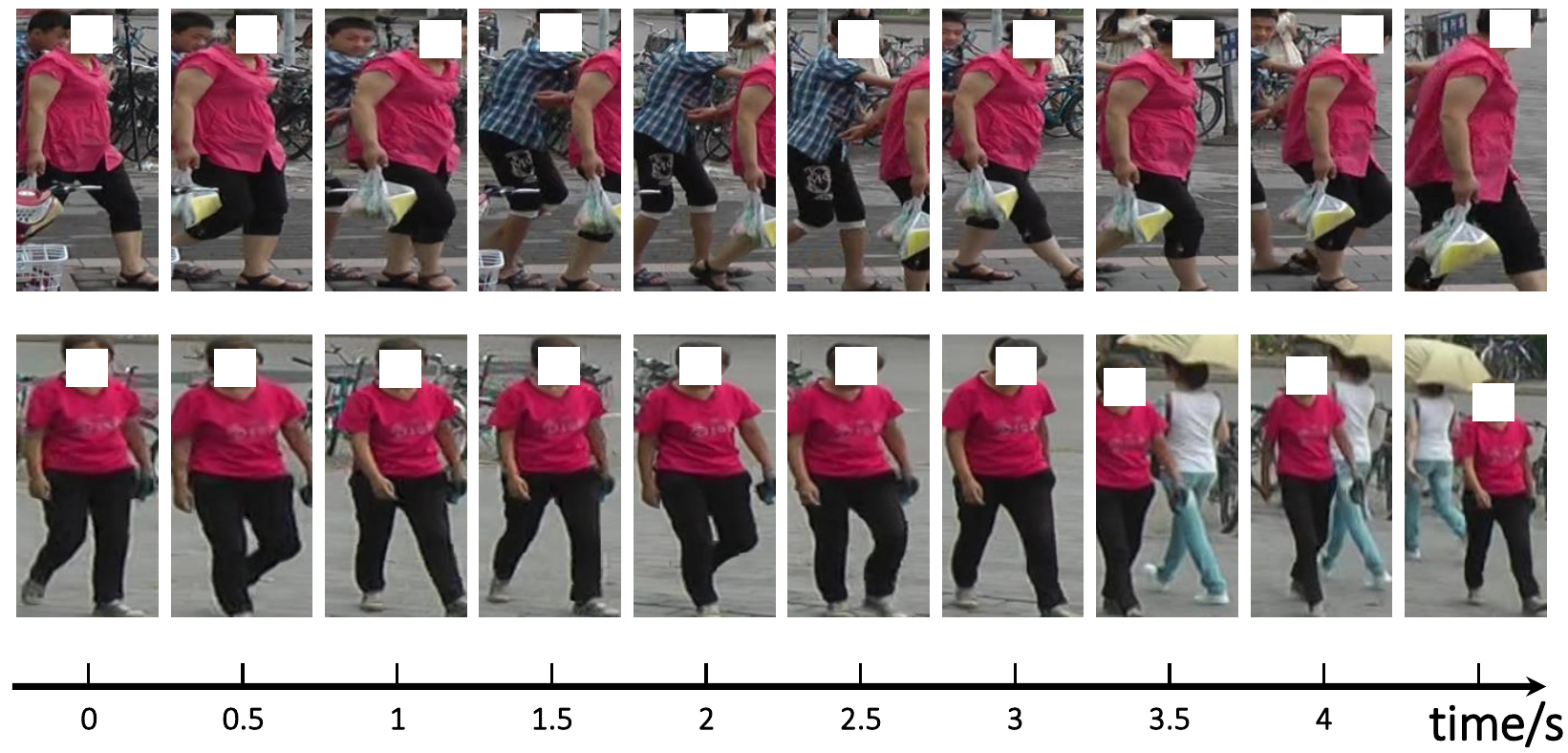}\\
\caption{Illustrations of two video sequences from two different pedestrians with similar appearance on MARS dataset (we cover the face for privacy purpose). Local temporal cues among adjacent frames, \emph{e.g.}, motion pattern or speed helps to differentiate those two pedestrians. The global contextual cues among adjacent frames can be applied to spot occlusions and noises, \emph{e.g.}, occluded frames show smaller similarity to other frames.}
\label{fig:sample}
\end{figure}

The key focus of existing studies for video person ReID lies on the exploitation of temporal cues.
Existing works can be divided into three categories according to their ways of temporal feature learning:
(i) extracting dynamic features from additional CNN inputs, \emph{e.g.}, through optical flow~\cite{mclaughlin2016recurrent,chung2017two};
(ii) extracting spatial-temporal features by regarding videos as $3$-dimensional data, \emph{e.g.}, through $3$D CNN~\cite{liu2019dense,li2019multi}.
(iii) learning robust person representations by temporally aggregating frame-level features, \emph{e.g.}, through Recurrent Neural Networks (RNN)~\cite{yan2016person,mclaughlin2016recurrent,chung2017two}, and temporal pooling or weight learning~\cite{liu2017video,zhou2017see,li2018diversity};

The third category, which our work belongs to, is currently dominant in video person ReID. The third category exhibits two advantages: (i) person representation techniques developed for image ReID can be easily explored compared to the first category; (ii) it avoids the estimation of optical flows, which is still not reliable enough due to misalignment errors between adjacent frames. Current studies have significantly boosted the performance on existing datasets, however they still show certain limitations in the aspects of either efficiency or the capability of temporal cues modeling. For instance, RNN model is complicated to train for long sequence videos. Feature temporal pooling could not model the order of video frames, which also conveys critical temporal cues. It is appealing to explore more efficient and effective way of acquiring spatial-temporal feature through end-to-end CNN learning.

This work targets to learn a discriminative Global-Local Temporal Representation (GLTR) from a sequence of frame features by embedding both short and long-term temporal cues. As shown in Fig.~\ref{fig:sample}, the short-term temporal cue among adjacent frames helps to distinguish visually similar pedestrians. The long-term temporal cue helps to alleviate the occlusions and noises in video sequences. Dilated spatial pyramid convolution~\cite{chen2017deeplab,yang2018denseaspp} is commonly used in image segmentation tasks to exploit the spatial contexts. Inspired by its strong and efficient spatial context modeling capability, this work generalizes the dilated spatial pyramid convolution to Dilated Temporal Pyramid (DTP) convolution for local temporal context learning. To capture the global temporal cues, a Temporal Self-Attention (TSA) model is introduced to exploit the contextual relations among inconsecutive frames. DTP and TSA are applied on frame-level features to learn the GLTR through end-to-end CNN training. As shown in our experiments and visualizations, GLTR presents strong discriminative power and robustness.

We test our approach on a newly proposed Large-Scale Video dataset for person ReID (LS-VID) and four widely used video ReID datasets, including PRID~\cite{hirzer2011person}, iLIDS-VID~\cite{wang2014person}, MARS~\cite{zheng2016mars}, and DukeMTMC-VideoReID~\cite{wu2018exploit,ristani2016performance}, respectively. Experimental results show that GLTR achieves consistent performance superiority on those datasets. It achieves Rank-1 Accuracy of $87.02\%$ on MARS without re-ranking, $2\%$ better than the recent PBR~\cite{suh2018part} that uses extra body part cues for video feature learning. It achieves Rank-1 Accuracy of $94.48\%$ on PRID and $96.29\%$ on DukeMTMC-VideoReID, which also beat the ones achieved by current state-of-the art.

GLTR representation is extracted by simple DTP and TSA models posted on a sequence of frame features. Although simple and efficient to compute, this solution outperforms many recent works that use complicated designs like body part detection and multi-stream CNNs. To our best knowledge, this is an early effort that jointly leverages dilated convolution and self-attention for multi-scale temporal feature learning in video person ReID.

\begin{figure*}
\centering
\includegraphics[width=0.95\linewidth]{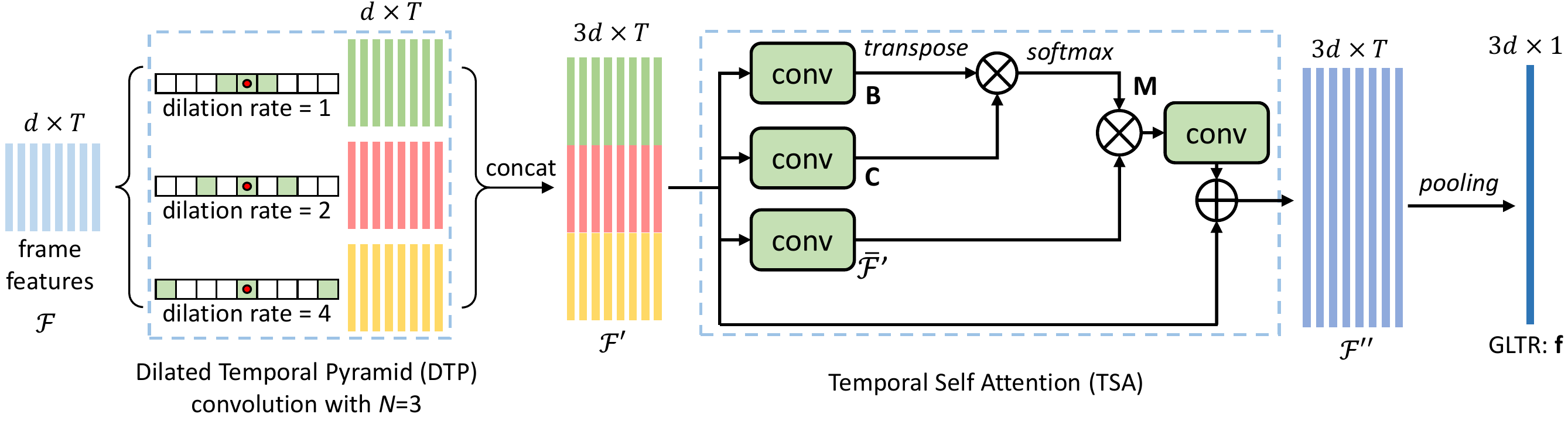}\\
\caption{Illustration of our frame feature aggregation subnetwork for GLTR extraction, which consists of Dilated Temporal Pyramid (DTP) convolution for local temporal context learning and Temporal Self-Attention (TSA) model to exploit the global temporal cues. }
\label{fig:framework}
\end{figure*}

\section{Related Work}

Existing person ReID works can be summarized into image based ReID~\cite{wang2014person,su2017pose, pedagadi2013local,xiong2014person,zhao2017deeply} and video based ReID~\cite{zheng2016mars,si2018dual,suh2018part,li2019multi}, respectively. This part briefly reviews four categories of temporal feature learning in video person ReID, which are closely related with this work.

\emph{Temporal pooling} is widely used to aggregate features across all time stamps.
Zheng \etal~\cite{zheng2016mars} apply max and mean pooling to get the video feature. Li \etal \cite{li2018diversity} utilize part cues and learn a weighting strategy to fuse features extracted from video frames. Suh \etal~\cite{suh2018part} propose a two-stream architecture to jointly learn the appearance feature and part feature, and fuse the image level features through a pooling strategy. Average pooling is also used in recent works~\cite{li2018unsupervised,wu2018exploit}, which apply unsupervised learning for video person ReID. Temporal pooling exhibits promising efficiency, but extracts frame features independently and ignores the temporal orders among adjacent frames.

\emph{Optical flow} encodes the short-term motion between adjacent frames. Many works utilize optical flow to learn temporal features~\cite{simonyan2014two,feichtenhofer2016convolutional,chung2017two}.
Simonyan \etal~\cite{simonyan2014two} introduce a two-stream network to learn spatial feature and temporal feature from stacked optical flows. Feichtenhofer \etal~\cite{feichtenhofer2017spatiotemporal} leverage optical flow to learn spatial-temporal features, and evaluate different types of motion interactions between two streams. Chung \etal~\cite{chung2017two} introduce a two stream architecture for appearance and optical flow, and investigate the weighting strategy for those two streams. Mclaughlin \etal~\cite{mclaughlin2016recurrent} introduce optical flow and RNN to exploit long and short term temporal cues. One potential issue of optical flow is its sensitive to spatial misalignment errors, which commonly exist between adjacent person bounding boxes.

\emph{Recurrent Neural Network} (RNN) is also adopted for video feature learning in video person ReID. Mclaughlin \etal~\cite{mclaughlin2016recurrent} first extract image level features, then introduce RNN to model temporal cues cross frames. The outputs of RNN are then combined through temporal pooling as the final video feature. Liu \etal~\cite{liu2019spatial}
propose a recurrent architecture to aggregate the frame-level representations and yield a sequence-level human feature representation. RNN introduces a certain number of fully-connected layers and gates for temporal cue modeling, making it complicated and difficult to train.

\emph{$3$D convolution} directly extracts spatial-temporal features through end-to-end CNN training. Recently, deep 3D CNN is introduced for video representation learning.
Tran \etal~\cite{tran2015learning} propose C3D networks for spatial-temporal feature learning.
Qiu \etal~\cite{qiu2017learning} factorize the 3D convolutional filters into spatial and temporal components, which yield performance gains.
Li \etal ~\cite{li2019multi} build a compact Multi-scale 3D (M3D) convolution network to learn multi-scale temporal cues. Although 3D CNN has exhibited promising performance, it is still sensitive to spatial misalignments and needs to stack a certain number of 3D convolutional kernels, resulting in large parameter overheads and increased difficult for CNN optimization.


This paper learns GLTR through posting DTP and TSA modules on frame features. Compared with existing temporal pooling strategies, our approach jointly captures global and local temporal cues, hence exhibits stronger temporal cue modeling capability. It is easier to optimize than RNN and presents better robustness to misalignment errors than optical flow. Compared with 3D CNN, our model has a more simple architecture and could easily leverage representations developed for image person ReID. As shown in our experiments, our approach outperforms the recent 3D CNN model M3D~\cite{li2019multi} and the recurrent model STMP~\cite{liu2019spatial}.

\section{Proposed Methods}

\subsection{Formulation}
Video person ReID aims to identify a gallery video that is about the same person with a query video from a gallery set containing $\mathcal{K}$ videos. A gallery video is denoted by $\mathcal{G}^k = \{{I}^k_{1}, {I}^k_{2},..., {I}^k_{T^k}\}$ with $k\in \{1, 2, ..., \mathcal{K}\}$, and the query video is denoted by $\mathcal{Q} = \{{I}^q_{1}, {I}^q_{2},..., {I}^q_{T^q}\}$, where $T^k$ ($T^q$) denotes the number of frames in the sequence and ${I}^k_{t}$ (${I}^q_{t}$) is the $t$-th frame. A gallery video $\mathcal{G}$ will be identified as true positive, if it has the closest distance to the query based on a video representation, \emph{i.e.},
\vspace{2mm}
\begin{align}
    \mathcal{G} = \arg\min_{k} \operatorname{dist}
    (\mathbf{f}^{\mathcal{G}^k}, \mathbf{f}^{\mathcal{Q}}),
\end{align}
where $\mathbf{f}^{\mathcal{G}^k}$ and $\mathbf{f}^{\mathcal{Q}}$ are the representations of the gallery video $\mathcal{G}^k$ and the query video $\mathcal{Q}$, respectively.

Our approach consists of two subnetworks to learn a discriminative video representation $\mathbf{f}$, \emph{i.e.}, image feature extraction subnetwork and frame feature aggregation subnetwork, respectively. The first subnetwork extracts features of $T$ frames, \emph{i.e.}, $\mathcal{F}=\{{f}_1, {f}_2, \dots, {f}_T\}$, where $f_t\in \mathcal{R}^d$. The second subnetwork aggregates the $T$ frame features into a single video representation vector. We illustrate the second subnetwork, which is the focus of this work in Fig.~\ref{fig:framework}. We briefly demonstrate the computation of DTP and TSA in the following paragraphs.

The DTP is designed to capture the local temporal cues among adjacent frames. As shown in Fig.~\ref{fig:framework}, DTP takes frame features in $\mathcal{F}$ as input and outputs the updated frame feature $\mathcal{F}' = \{f'_1, f'_2, \dots, f'_T\}$. Each $f'_t \in \mathcal{F}'$ is computed by aggregating its adjacent frame features, \emph{i.e.},
\begin{equation}
\begin{aligned}
f'_t =\mathcal{M}_{DTP}(f_{t-i},...,f_{t+i}),
\end{aligned}
\label{equ:local}
\end{equation}
where $\mathcal{M}_{DTP}$ denotes the DTP model, and $f'_t$ is computed from $2\times i$ adjacent frames.

The TSA model exploits the relation among inconsecutive frames to capture the global temporal cues. It takes $\mathcal{F}'$ as input and outputs the temporal feature $\mathcal{F}'' = \{f''_1, f''_2, \dots, f''_T\}$. Each $f''_t \in \mathcal{F}''$ is computed by considering the contextual relations among features inside $\mathcal{F}'$, \emph{i.e.},
\begin{equation}
\begin{aligned}
\ f''_t = \mathcal{M}_{TSA}(\mathcal{F}', f'_t),
\end{aligned}
\label{equ:local}
\end{equation}
where $\mathcal{M}_{TSA}$ is the TSA model.

Each $f''_t$ aggregates both local and global temporal cues. We finally apply average pooling on $\mathcal{F}''$ to generate the fixed length GLTR $\mathbf{f}$ for video person ReID, \emph{i.e.},
\begin{align}
\mathbf{f} =\frac{1}{T}\sum^T_{t=1} f''_t.
\label{eq:avgpool}
\end{align}
Average pooling is also commonly applied in RNN~\cite{mclaughlin2016recurrent} and 3DCNN~\cite{li2019multi} to generate fixed-length video feature. The global and local temporal cues embedded in each $f''_t$ guarantee the strong discriminative power and robustness of $\mathbf{f}$. The following parts introduce the design of DTP and TSA.

\subsection{Dilated Temporal Pyramid Convolution}\label{sec:ltr}

\noindent\textbf{Dilated Temporal Convolution:}
Dilated spatial convolution has been widely used in image segmentation for its efficient spatial context modeling capability~\cite{yu2015multi}. Inspired by dilated spatial convolution, we implement dilated temporal convolution for local temporal feature learning. Suppose the ${W}\in \mathcal{R}^{d\times w}$ is a convolutional kernel with temporal width $w$. With input frame features $\mathcal{F} = \{{f}_1, {f}_2, \dots, {f}_T\}$, the output $\mathcal{F}^{(r)}$ of dilated convolution with dilation rate $r$ can be defined as,
\begin{equation}
\begin{aligned}
\mathcal{F}^{(r)} = \{\ f_1^{(r)}, f_2^{(r)}, \ ..., \ f_T^{(r)}\}, \\
f_t^{(r)}= \sum _{i=1}^{w}f_{[t+r\cdot i]}\times {W}_{[i]}^{(r)}, f_t^{(r)} \in \mathcal{R}^{d},
\end{aligned}
\label{equ:mutual}
\end{equation}
where $\mathcal{F}^{(r)}$ is the collection of output features containing $f^{(r)}_t$. ${W}^{(r)}$ denotes dilated convolution with dilation rate $r$.

The dilation rate $r$ indicates the temporal stride for sampling frame features. It decides the temporal scales covered by dilated temporal convolution. For instance, with $r=2, w=3$, each output feature corresponds to a temporal range of five adjacent frames. Standard convolution can be regarded as a special case with $r=1$, which covers three adjacent frames. Compared with standard convolution, dilated temporal convolution with $r \geq 2$ has the same number of parameters to learn, but enlarges the receptive field of neurons without reducing the temporal resolution. This property makes dilated temporal convolution an efficient strategy for multi-scale temporal feature learning.

\vspace{.1cm}
\noindent\textbf{Dilated Temporal Pyramid Convolution:}
Dilated temporal convolutions with different dilation rates model temporal cues at different scales. We hence use parallel dilated convolutions to build the DTP convolution to enhance its local temporal cues modeling ability.

As illustrated in Fig.~\ref{fig:framework}, DTP convolution consists of $N$ parallel dilated convolutions with dilation rates increasing progressively to cover various temporal ranges. For $n$-th dilated temporal convolution, we set its dilation rate $r_n$ as $r_n = 2^{n-1}$ to efficiently enlarge the temporal receptive fields. We concatenate the outputs from $N$ branches as the updated temporal feature $\mathcal{F}'$, \emph{i.e.}, we compute $f'_t \in \mathcal{F}'$ as
\begin{align}\label{eq:atp_rep}
f'_t = concat(f^{(r_1)}_t,\ f^{(r_2)}_t,...,\ f^{(r_N)}_t ), f'_t \in \mathcal{R}^{Nd},
\end{align}
where $r_i$ is the dilation rate of $i$-th dilated temporal convolutions.

\begin{table*}
\small
\tabcolsep=3.5pt
\centering
\caption{The statistics of our {LS-VID} dataset and other video person ReID datasets.}
\vspace{1mm}
\label{table:datasets}%
\begin{tabular}{l|l|l|l|c|c|c|c|c|c}
\hline
dataset  &\#identity &\#sequence &\#boxes &\#frame &\#indoor cam.   &
\#outdoor cam. &detector & val. set &evaluation \\
\hline
{DukeMTMC}  &1,404   &4,832   &815,420    &168   &0    &8  &Hand        & $\times$  &CMC + mAP\\

{MARS}                &1,261   &20,715  &1,067,516  &58    &0    &6  &DPM     &  $\times$ &CMC + mAP\\

{PRID}                &200     &400     &40,033     &100   &0    &2  &Hand      & $\times$ &CMC\\

{iLIDS-VID}           &300     &600     &42,460     &73    &2    &0  &Hand        & $\times$ &CMC\\
{\textbf{LS-VID}}     &\textbf{3,772}   &\textbf{14,943}  &\textbf{2,982,685}  &\textbf{200}   &\textbf{3}    &\textbf{12} &\textbf{Faster R-CNN} &  \checkmark &\textbf{CMC + mAP}\\
\hline
\end{tabular}
\end{table*}

\begin{figure}
\centering
\includegraphics[width=1\linewidth]{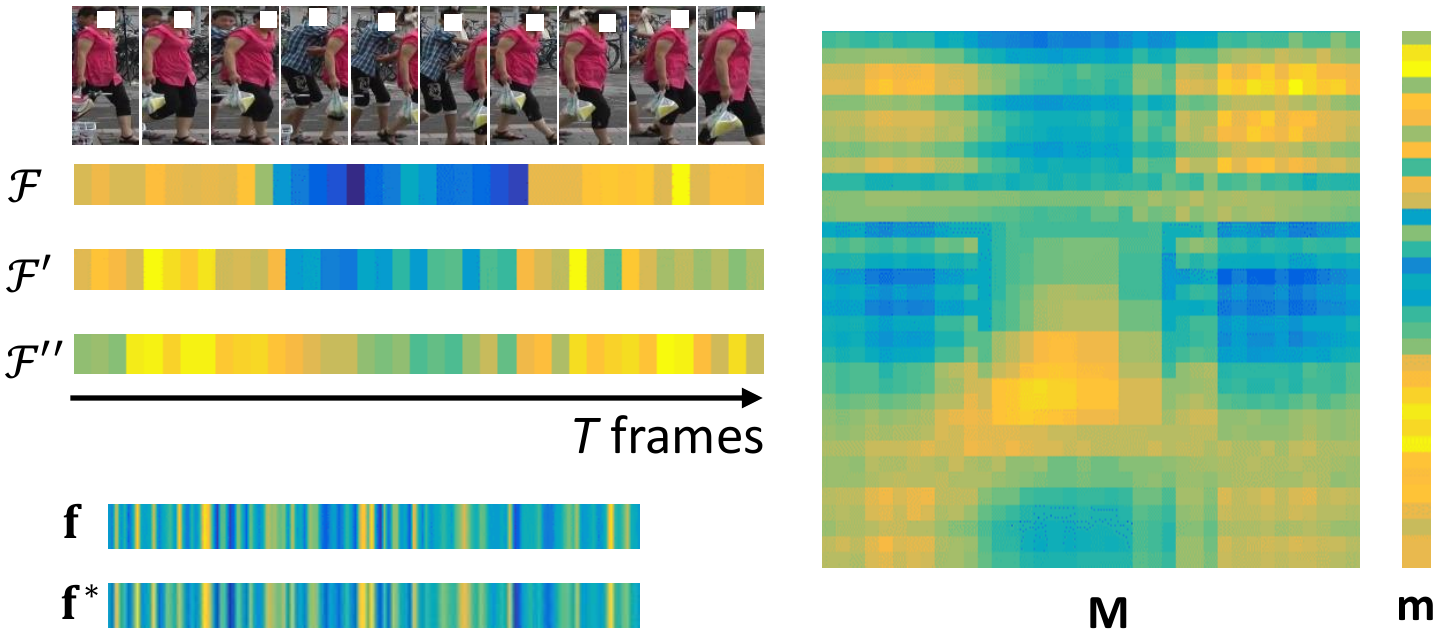}\\
\caption{Visualization of $\mathcal{F}$, $\mathcal{F'}$, $\mathcal{F''}$, $\textbf{M}$ and $\textbf{f}$ computed on a tracklet with occlusions. Dimensionality of $\mathcal{F}$, $\mathcal{F'}$, $\mathcal{F''}$ is reduced to $1\times T$ by PCA for visualization. It is clear that, occlusion affects the baseline feature $\mathcal{F}$, \emph{i.e.}, feature substantially changes as occlusion happens. DTP and TSA progressively alleviate the occlusions,\emph{ i.e.}, features of occluded frames in $\mathcal{F}'$ and $\mathcal{F}''$ appear similar to the others. $\textbf{f}^*$ is generated after manually removing occluded frames. $\textbf{f}$ is quite close to $\textbf{f}^*$, indicating the strong robustness of GLTR to occlusion. }
\label{fig:visualize}
\end{figure}

\subsection{Temporal Self Attention}

\vspace{.1cm}
\noindent\textbf{Self-Attention:} The self-attention module is recently used to learn the long-range spatial dependencies in image segmentation~\cite{fu2019dual,huang2018ccnet,yuan2018ocnet}, action recognition~\cite{wang2018non} and image person ReID~\cite{jiangself,ainam2018self}. Inspired by its promising performance in spatial context modeling, we generalize self-attention to to capture the contextual temporal relations among inconsecutive frames.

\noindent\textbf{Temporal Self-Attention:} The basic idea of TSA is to compute an $T\times T$ sized attention mask $\textbf{M}$ to store the contextual relations among all frame features. As illustrated in Fig.~\ref{fig:framework}, given the input $\mathcal{F}' \in R^{\ Nd\times T}$, TSA first applies two convolution layers followed by Batch Normalization and ReLU to generate feature maps $\textbf{B}$ and $\textbf{C}$ with size $(Nd/\alpha)\times T$, respectively. Then, it performs a matrix multiplication between $\textbf{C}$ and the transpose of $\textbf{B}$, resulting in a ${T\times T}$ sized temporal attention mask $\textbf{M}$.

$\textbf{M}$ is applied to update the $\mathcal{F}'$ to embed extra global temporal cues. $\mathcal{F}'$ is fed into a convolution layer to generate a new feature map $\bar{\mathcal{F'}}$ with size $(Nd/\alpha)\times T$. $\bar{\mathcal{F'}}$ is hence multiplied by $\textbf{M}$ and is fed into a convolution layer to recover its size to $Nd\times T$. The resulting feature map is fused with the original $\mathcal{F}'$ by residual connection, leading to the updated temporal feature $\mathcal{F}''$. The computation of TSA can be denoted as
\begin{equation}
\mathcal{F}'' = W*( \bar{\mathcal{F'}}\cdot\textbf{M})+\mathcal{F}',\mathcal{F}'' \in \mathcal{R}^{Nd \times T},
\label{equ:nonlocal2}
\end{equation}
where $W$ denotes the last convolutional kernel. $W$ is initialized as 0 to simplify the optimization of residual connection. $\alpha$ controls the parameter size in TSA. We experimentally set $\alpha$ as 2. $\mathcal{F}''$ is processed with average pooling to generate the final GLTR $\textbf{f} \in \mathcal{R}^{Nd}$.

We visualize the $\mathcal{F}$, $\mathcal{F'}$, $\mathcal{F''}$, $\textbf{M}$, and $\textbf{f}$ computed on a tracklet with occlusion in Fig.~\ref{fig:visualize}. DTP reasonably alleviates occlusion by applying convolutions to adjacent features. TSA alleviates occlusion mainly by computing the attention mask $\textbf{M}$, which stores the global contextual relations as shown in Fig.~\ref{fig:visualize}. With $\textbf{M}$, average pooling on $\mathcal{F}''$ can be conceptually expressed as:
\begin{equation}
\begin{aligned}
\sum _{t=1}^{T}\mathcal{F}''(:,t) \doteq \sum _{t=1}^{T}\mathcal{F'}(:,t)\times \textbf{m}(t)+\sum _{t=1}^{T}\mathcal{F}'(:,t),\\
\end{aligned}
\label{eq:maskM}
\end{equation}
where $\textbf{m}=\sum _{t=1}^{T}\textbf{M}(:,t)$ is a $T$-dim weighting vector. Note that, Eq.~\eqref{eq:maskM} omits the convolutions before and after $\bar{\mathcal{F'}}$ to simplify the expression. $\textbf{m}$ is visualized in Fig.~\ref{fig:visualize}, where occluded frames presents lower weights, indicating their features are depressed during average pooling. Combining DTP and TSA, GLTR presents strong robustness.

\begin{figure}
\centering
\includegraphics[width=1\linewidth]{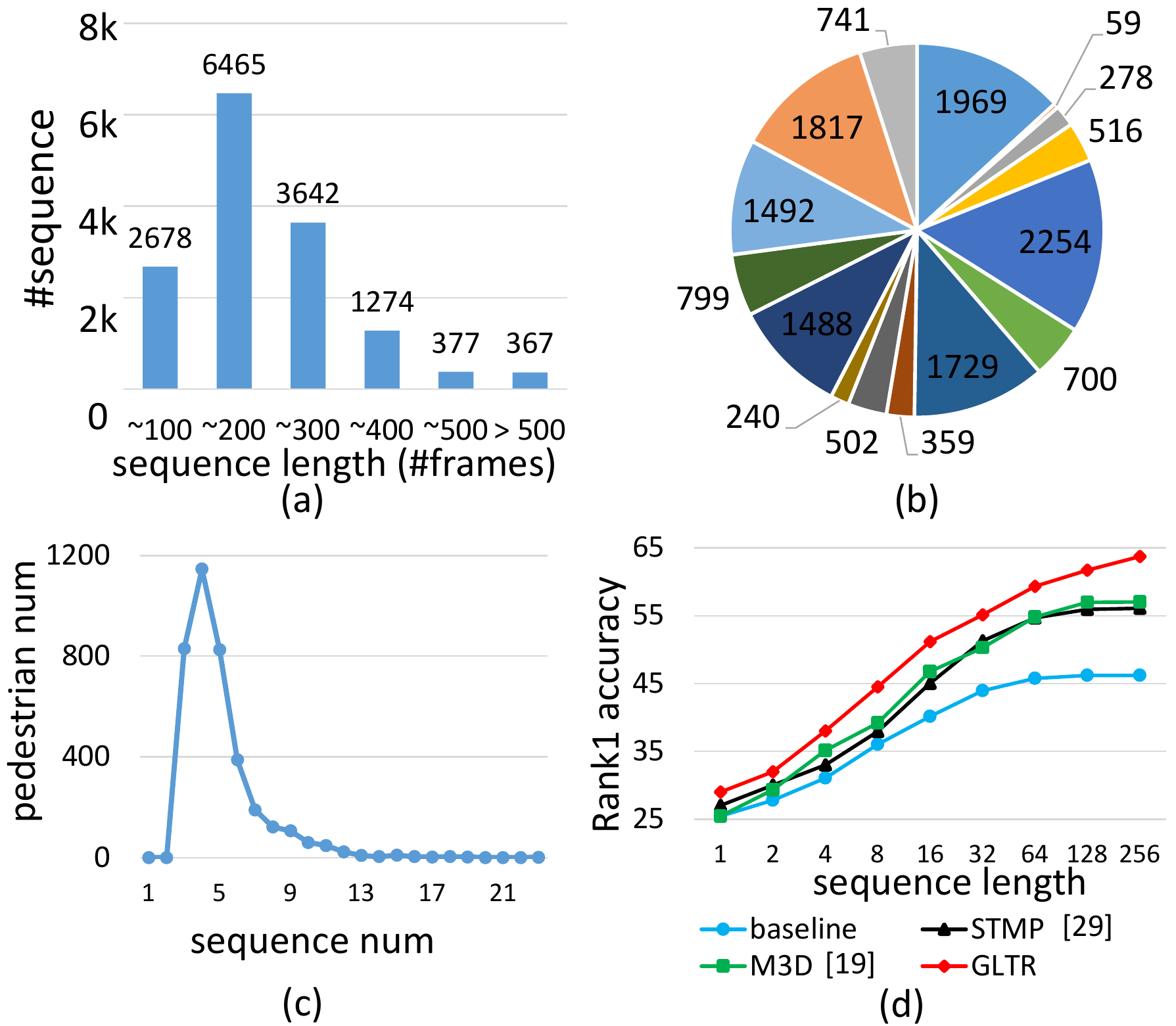}
\caption{Some statistics on {LS-VID} dataset: (a) the number of sequences with different length; (b) the number of sequences in each of the 15 cameras; (c) the number of identities with different sequence number; (d) the ReID performance with different testing sequence length.}
\label{fig:ourstatistic}
\end{figure}

\begin{figure}
\centering
\includegraphics[width=1\linewidth]{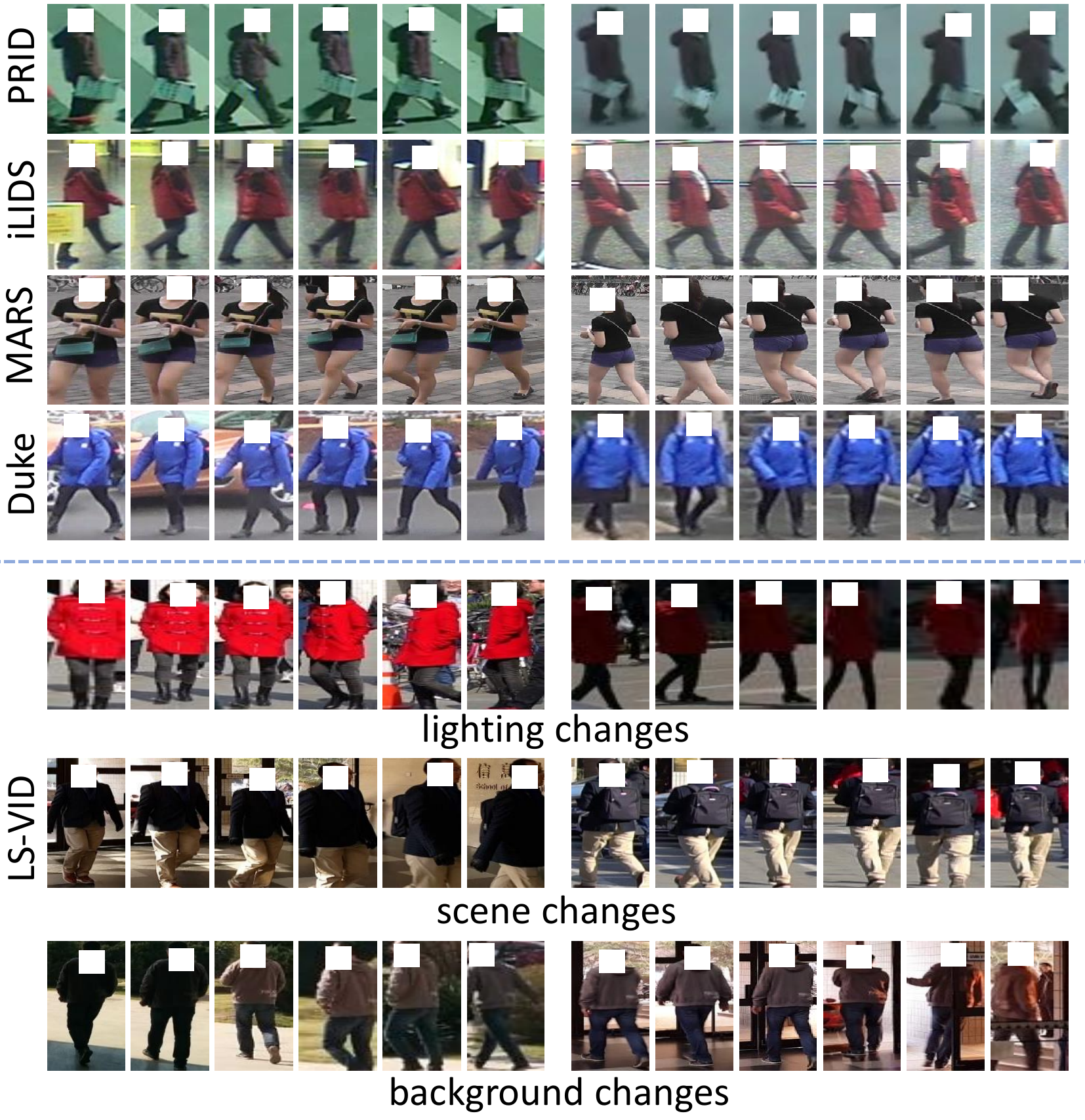}\\
\caption{ Frames evenly sampled from person tracklets. Each row shows two sequences of the same person under different cameras. Compared with existing datasets, {LS-VID} presents more substantial variations of lighting, scene, and background, \emph{etc}. We cover the face in for privacy purpose.}
\label{fig:datasample}
\end{figure}

\section{Experiment}
\subsection{Dataset}

We test our methods on four widely used video ReID datasets, and a novel large-scale dataset. Example images are depicted in Fig.~\ref{fig:datasample} and statistics are given in Table~\ref{table:datasets}.

\vspace{.1cm}
\noindent
\textbf{PRID-2011~\cite{hirzer2011person}.}
There are 400 sequences of 200 pedestrians captured by two cameras. Each sequence has a length between 5 and 675 frames.

\vspace{.1cm}
\noindent
\textbf{iLIDS-VID}~\cite{wang2014person}.
There are 600 sequences of 300 pedestrians from two cameras. Each sequence has a variable length between 23 and 192 frames. Following the implementation in previous works~\cite{wang2014person,li2018diversity}, we randomly split this two datasets into train/test identities. This procedure is repeated 10 times for computing averaged accuracies.

\vspace{.1cm}
\noindent
\textbf{MARS}~\cite{zheng2016mars}. This dataset is captured by 6 cameras. It consists of 17,503 sequences of 1,261 identities and 3,248 distractor sequences. It is split into 625 identities for training and 636 identities for testing. The bounding boxes are detected with DPM detector~\cite{felzenszwalb2010object}, and tracked using the GMMCP tracker~\cite{dehghan2015gmmcp}. we follow the protocol of MARS and report the Rank1 accuracy and mean Average Precision (mAP).

\vspace{.1cm}
\noindent
\textbf{DukeMTMC-VideoReID}~\cite{wu2018exploit,ristani2016performance}. There are 702 identities for training, 702 identities for testing, and 408 identities as distractors. The training set contains 369,656 frames of 2,196 tracklets, and test set contains 445,764 frames of 2,636 tracklets.

\vspace{.1cm}
\noindent
\textbf{{LS-VID}.}
Besides the above four datasets, we collect a novel Large-Scale Video dataset for person ReID (LS-VID).

\emph{Raw video capture:} We utilize a 15-camera network and select 4 days for data recording. For each day, 3 hours of videos are taken in the morning, noon, and afternoon, respectively. Our final raw video contains 180 hours videos, 12 outdoor cameras, 3 indoor cameras, and 12 time slots.

\emph{Detection and tracking:}
Faster RCNN~\cite{ren2015faster} is utilized for pedestrian detection. After that, we design a feature matching strategy to track each detected pedestrian in each camera. After discarding some sequences with too short length, we finally collect 14,943 sequences of 3,772 pedestrians, and the average sequence length is 200 frames.

\emph{Characteristics:}
Example sequences in LS-VID are shown in Fig.~\ref{fig:datasample}, and statistics are given in Table~\ref{table:datasets} and Fig.~\ref{fig:ourstatistic}. LS-VID shows the following new features: (1)Longer sequences. (2) More accurate pedestrian tracklets. (3) Currently the largest video ReID dataset. (4) Define a more realistic and challenging ReID task.

\emph{Evaluation protocol:}
Because of the expensive data annotation, we randomly divide our dataset into training set and test set with 1:3 ratio to encourage more efficient training strategies. We further divide a small validation set. Finally, the training set contains 550,419 bounding boxes of 842 identities, the validation set contains 155,191 bounding boxes of 200 identities, and the test set contains 2,277,075 bounding boxes of 2,730 identities. Similar to existing video ReID datasets~\cite{zheng2016mars,wu2018exploit}, LS-VID utilizes the Cumulated Matching Characteristics (CMC) curve and mean Average Precision (mAP) as evaluation metric.

\begin{figure}
\centering
\includegraphics[width=1\linewidth]{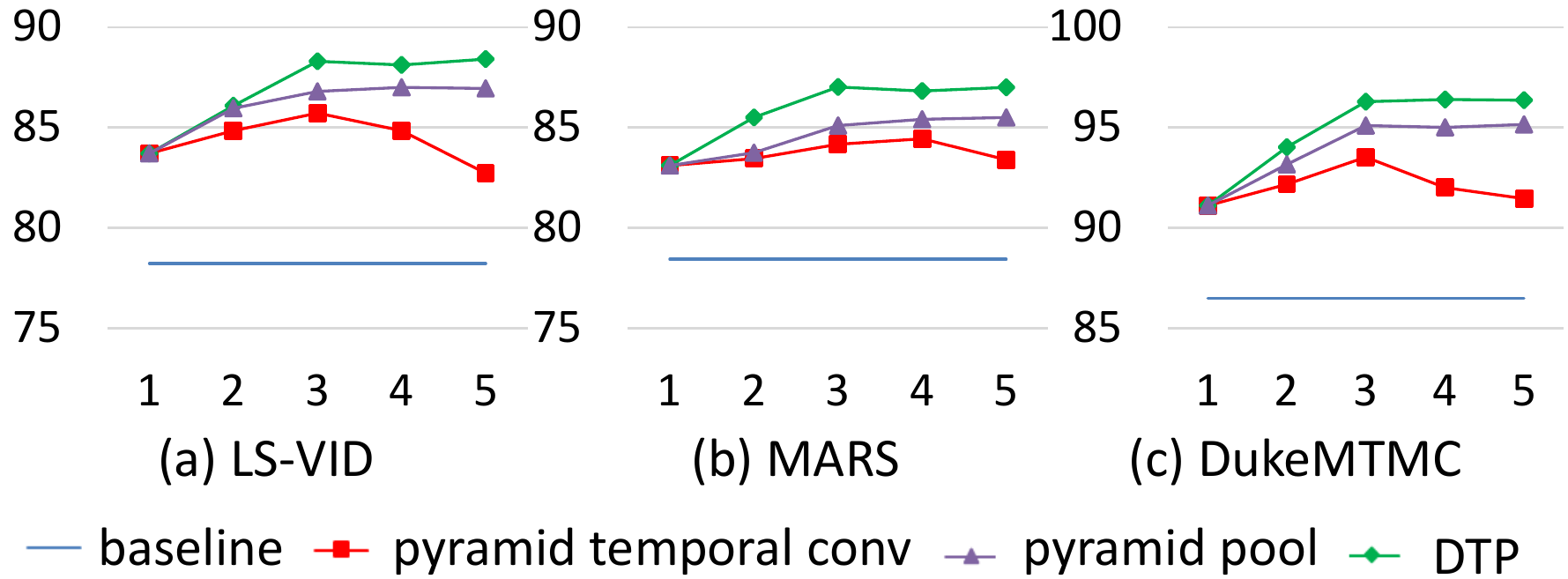}\\
\caption{Rank1 accuracy of DTP and two competitors on three datasets with different numbers of branches, \emph{i.e.}, parameter $N$.}
\label{fig:ltrcompare}
\end{figure}

\subsection{Implementation Details}
We employ standard ResNet50~\cite{he2016deep} as the backbone for frame feature extraction. All models are trained and fine-tuned with PyTorch. Stochastic Gradient Descent (SGD) is used to optimize our model. Input images are resized to 256$\times$128. The mean value is subtracted from each (B, G, and R) channel. For 2D CNN training, each batch contains 128 images. The initial learning rate is set as 0.01, and is reduced ten times after 10 epoches. The training is finished after 20 epoches. For DTP and TSA training, we sample $16$ adjacent frames from each sequence as input for each training epoch. The batch size is set as 10. The initial learning rate is set as 0.01, and is reduced ten times after 120 epoches. The training is finished after 400 epoches. All models are trained with only softmax loss.

During testing, we use 2D CNN to extract a $d$=128-dim feature from each video frame, then fuse frame features into GLTR using the network illustrated in Fig.~\ref{fig:framework}. The video feature is finally used for person ReID with Euclidean distance. All of our experiments are implemented with GTX TITAN X GPU, Intel i7 CPU, and 128GB memory.

\subsection{Ablation Study}

\noindent\textbf{Comparison of DTP and other local temporal cues learning strategies:} Besides DTP, we also implement the following strategies to learn temporal cues among adjacent frames: (i) pyramid temporal convolution without dilation, and (ii) temporal pyramid pooling~\cite{zhao2017pyramid}. As explained in Sec.~\ref{sec:ltr}, the dilation rate of $i$-th pyramid branch in DTP is $r_i=2^{i-1}$. To make a fair comparison, we set three methods have the same number of branches, where each has the same size of receptive field. For instance, we set the convolution kernel size as $d\times 9$ for the 3-rd branch of pyramid temporal convolution without dilation. The experiment results on MARS, DukeMTMC-VideoReID, and the \textbf{validation} set of LS-VID are summarized in Fig.~\ref{fig:ltrcompare}.

Fig.~\ref{fig:ltrcompare} also compares average pooling as the baseline. It is clear that, three methods perform substantially better than baseline, indicating that average pooling is not effective in capturing the temporal cues among frame features. With N=1, the three methods perform equally, \emph{i.e.}, apply a $d\times 3$ sized convolution kernel to frame feature $\mathcal{F}$. As we increase $N$, the performances of three algorithms are boosted. This means that introducing multiple convolution scales benefits the learned temporal feature.

It is also clear that, DTP consistently outperforms the other two strategies on three datasets. The reason may be because the temporal pyramid pooling loses certain temporal cues as it down-samples the temporal resolution. The traditional temporal convolution introduces too many parameters, leading to difficult optimization. The dilated convolutions in DTP efficiently enlarge the temporal respective fields hence performs better for local temporal feature learning. With $N\geq 3$, the performance boost slows down for DTP. Further introducing more branches increases the size of parameters and causes more difficult optimization. We select $N=3$ for DTP in the following experiments.

\begin{table}
\footnotesize
\begin{center}
\caption{Performance of individual components in GLTR.}
\vspace{1mm}
\label{table:gltr}
\setlength{\tabcolsep}{3.3 pt}
\begin{tabular}{l|c|c|c|c|c|c|c|c}
\hline
Dataset  &\multicolumn{2}{c|}{LS-VID}  &\multicolumn{2}{c|}{MARS} &\multicolumn{2}{c|}{DukeMTMC} &PRID &iLIDS\\
\hline
Method  &mAP&rank1      &mAP&rank1  &mAP&rank1  &rank1 &rank1\\
\hline
baseline &30.72&46.18  &65.45&78.43 &82.08&86.47 &83.15 &62.67\\
\hline
DTP      &41.78&59.92  &75.90&85.74 &89.98&93.02 &93.26&84.00\\
TSA      &40.01&58.73  &75.62&85.40 &89.26&92.74 &92.14&83.33\\
\hline
GLTR     &\textbf{44.32}&\textbf{63.07}  &\textbf{78.47}&\textbf{87.02} &\textbf{93.74}&\textbf{96.29}  &\textbf{95.50}&\textbf{86.00}\\
\hline
\end{tabular}
\end{center}
\vspace{-1mm}
\end{table}

\begin{table}
\footnotesize
\begin{center}
\caption{Performance of GLTR with different backbones on {LS-VID} test set.}
\vspace{1mm}
\label{table:backbone}
\setlength{\tabcolsep}{4.5pt}
\begin{tabular}{l|l|c|c|c|c|c}
\hline
method    &backbone  &mAP&rank1&rank5&rank10&rank20\\
\hline
\multirow{3}{*}{baseline}
&Alexnet~\cite{krizhevsky2012imagenet} &15.98&24.23&43.52&53.45&62.13\\\cline{3-7}
&Inception~\cite{szegedy2016rethinking}&22.77&35.70&55.88&64.89&73.12\\\cline{3-7}
&ResNet50~\cite{he2016deep}            &30.72&46.18&67.41&74.71&82.33\\\cline{3-7}
\hline
\multirow{3}{*}{GLTR}
&Alexnet~\cite{krizhevsky2012imagenet} &22.57&35.45&56.59&66.01&75.06\\\cline{3-7}
&Inception~\cite{szegedy2016rethinking}&35.75&51.83&71.66&79.19&84.79\\\cline{3-7}
&ResNet50~\cite{he2016deep}            &\textbf{44.43}&\textbf{63.07}&\textbf{77.22}&\textbf{83.81}&\textbf{88.41}\\\cline{3-7}
\hline
\end{tabular}
\end{center}
\vspace{-3mm}
\end{table}
\noindent\textbf{Validity of combining DTP and TSA:} This part proceeds to evaluate that combining DTP and TSA results in the best video feature. We compare several variants of our methods and summarize the results on four datasets and the \textbf{test set} of LS-VID in Table~\ref{table:gltr}. In the table, ``baseline" denotes the ResNet50 + average pooling. ``DTP" and ``TSA" denote aggregating frame feature only with DTP or TSA, respectively. ``GLTR" combines DTP and TSA.

Table~\ref{table:gltr} shows that either DTP or TSA performs substantially better than the baseline, indicating modeling extra local and global temporal cues results in better video feature. DTP model achieves rank1 accuracy of 85.74\% on MARS dataset, outperforming the baseline by large margin. Similarly, TSA also performs substantially better than the baseline. By combining DTP and TSA, the GLTR consistently achieves the best performance on five datasets. We hence conclude that, jointly learning local and global temporal cues results in the best video feature.


\vspace{1mm}
\noindent\textbf{Different backbones:} We further evaluate the effectiveness of GLTR with different backbone networks, including Alexnet~\cite{krizhevsky2012imagenet}, Inception~\cite{szegedy2016rethinking} and ResNet50~\cite{he2016deep}. Experimental results on the test set of LS-VID are summarized in Table~\ref{table:backbone}. Table~\ref{table:backbone} shows that, implemented on different backbones, GLTR consistently outperforms baselines, indicating that our methods work well with different frame feature extractors. GLTR thus could leverage strong image representations and serve as a general solution for video person ReID. Since ResNet50 achieves best performance in Table~\ref{table:backbone}, we adopt ResNet50 in the following experiments.

\subsection{Comparison With Recent Works}

\begin{table}
\footnotesize
\caption{Comparison with recent works on LS-VID test set.}
\label{table:lsvid}
\setlength{\tabcolsep}{7pt}
\begin{center}
\begin{tabular}{l|c|c|c|c|c}
\hline
Method        &mAP  &rank1   &rank5   &rank10  &rank20\\
\hline
ResNet50~\cite{he2016deep}   &30.72&46.18&67.41&74.71&82.33\\
GLAD~\cite{wei2017glad}      &33.98&49.34&70.15&77.14&83.59\\
HACNN~\cite{li2018harmonious}&36.65&53.93&72.41&80.88&85.27\\
PBR~\cite{suh2018part}       &37.58&55.34&74.68&81.56&86.16\\
DRSA~\cite{li2018diversity}  &37.77&55.78&74.37&81.06&86.81\\
\hline
Two-stream~\cite{simonyan2014two}   &32.12&48.23&68.66&75.06&83.56\\
LSTM~\cite{yan2016person} &35.92&52.11&72.57&78.91&85.50\\
I3D~\cite{carreira2017quo}  &33.86&51.03&70.08&78.08&83.65\\
P3D~\cite{qiu2017learning}  &34.96&53.37&71.15&78.08&83.65\\
STMP~\cite{liu2019spatial} &39.14&56.78&76.18&82.02&87.12\\
M3D~\cite{li2019multi}     &40.07&57.68&76.09&83.35&88.18\\
\hline
GLTR  &\textbf{44.32}&\textbf{63.07}&\textbf{77.22}&\textbf{83.81}&\textbf{88.41}\\
\hline
\end{tabular}
\end{center}
\vspace{-4mm}
\end{table}

\noindent\textbf{LS-VID:}
This section compares several recent methods with our approach on LS-VID \textbf{test set}. To make a comparison on LS-VID, we implement several recent works with code provided by their authors, including temporal feature learning methods for person reid: M3D~\cite{li2019multi} and STMP~\cite{liu2019spatial}, other temporal feature learning methods: two-stream
CNN with appearance and optical flow~\cite{simonyan2014two}, LSTM~\cite{yan2016person}, 3D convolution: I3D~\cite{carreira2017quo} and P3D~\cite{qiu2017learning}, as well as recent person ReID works: GLAD~\cite{wei2017glad}, HACNN~\cite{li2018harmonious}, PBR~\cite{suh2018part} and DRSA~\cite{li2018diversity}, respectively. Video features of GLAD~\cite{wei2017glad} and HACNN~\cite{li2018harmonious} are extracted by average pooling. We repeat PBR~\cite{suh2018part} and DRSA~\cite{li2018diversity} by referring to their implantations on MARS. Table~\ref{table:lsvid} summarizes the comparison.

Table~\ref{table:lsvid} shows that, GLAD~\cite{wei2017glad} and HACNN~\cite{li2018harmonious} get promising performance in image person ReID, but achieve lower performance than temporal feature learning strategies, \eg, M3D~\cite{li2019multi} and STMP~\cite{liu2019spatial}. This indicates the importance of learning temporal cues in video person ReID. Among those compared temporal feature learning methods, the recent M3D achieves the best performance. In Table~\ref{table:lsvid}, the proposed GLTR achieves the best performance. It outperforms the recent video person ReID work STMP~\cite{liu2019spatial} and M3D~\cite{li2019multi} by large margins, \emph{e.g.}, $6.29\%$ and $5.39\%$ in rank1 accuracy, respectively.

\begin{table}
\footnotesize
\caption{Comparison with recent works on {MARS}.}
\label{table:comparemars}
\setlength{\tabcolsep}{10pt}
\begin{center}
\begin{tabular}{l|c|c|c|c}
\hline
Method&mAP&rank1&rank5&rank20 \\
\hline
BoW+kissme~\cite{zheng2016mars}   &15.50&30.60&46.20&59.20\\
IDE+XQDA~\cite{zheng2016mars}     &47.60&65.30&82.00&89.00\\
SeeForest~\cite{zhou2017see}      &50.70&70.60&90.00&97.60\\
QAN~\cite{liu2017quality}         &51.70&73.70&84.90&91.60\\
DCF~\cite{li2017learning}         &56.05&71.77&86.57&93.08\\
TriNet~\cite{hermans2017defense}  &67.70&79.80&91.36&-\\
MCA~\cite{song2018mask}           &71.17&77.17&-&-\\
DRSA~\cite{li2018diversity}       &65.80&82.30&-&-\\
DuATM~\cite{si2018dual}           &67.73&81.16&92.47&-\\
MGCAM~\cite{song2018mask}         &71.17&77.17&-&-\\
PBR~\cite{suh2018part}            &75.90&84.70&92.80&95.00\\
CSA~\cite{chen2018video}          &76.10&86.30&94.70&98.20\\
STMP~\cite{liu2019spatial}        &72.70&84.40&93.20&96.30\\
M3D~\cite{li2019multi}            &74.06&84.39&93.84&97.74\\
STA~\cite{fu2019sta}              &\textbf{80.80}&86.30&95.70&98.10\\
\hline
GLTR                              &78.47&\textbf{87.02}&\textbf{95.76}&\textbf{98.23}\\
\hline
\end{tabular}
\end{center}
\vspace{-5mm}
\end{table}

\begin{figure*}
\centering
\includegraphics[width=1\linewidth]{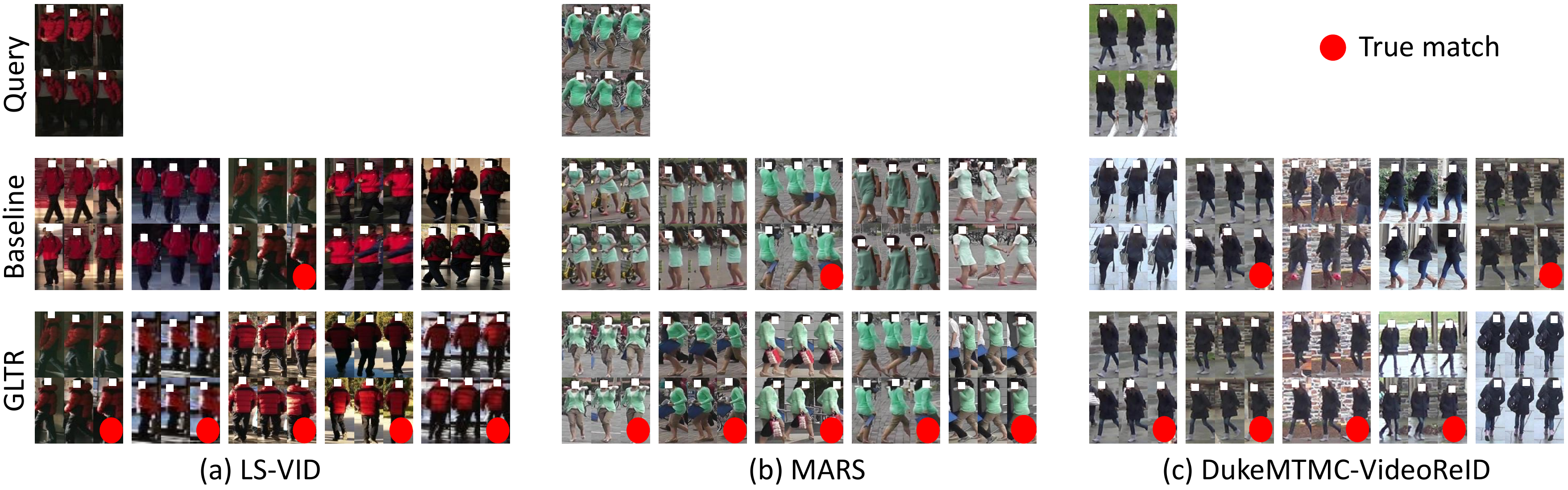}\\
\caption{Illustration of person ReID results on LS-VID, MARS and DukeMTMC-VideoReID datasets. Each example shows the top-5 retrieved sequences by baseline method (first tow) and GLTR (second tow), respectively. The true match is annotated by the red dot. We cover the face for privacy purpose.}
\label{fig:result}
\end{figure*}

\vspace{.1cm}
\noindent\textbf{MARS:}
Table~\ref{table:comparemars} reports the comparison with recent works on MARS. GLTR achieves the rank1 accuracy of 87.02\% and mAP of 78.47\%, outperforming most of the recent work, \eg, STMP~\cite{liu2019spatial}, M3D~\cite{li2019multi} and STA~\cite{fu2019sta} by 2.62\%, 2.63\%, and 0.72\% in rank1 accuracy, respectively.
Note that, STMP~\cite{liu2019spatial} introduces a complex recurrent network and uses part cues and triplet loss. M3D~\cite{li2019multi} use 3D CNN to learn the temporal cues, hence requires higher computational complexity.
STA~\cite{fu2019sta} achieves competitive performance on MARS dataset, and outperform GLTR on mAP. Note that, STA introduces multi-branches for part feature learning and uses triplet loss to promote the performance.
Compared with those works, our method achieves competitive performance with simple design., \emph{e.g.}, we extract global feature with basic backbone and train only with the softmax loss. GLTR can be further combined with a re-ranking strategy~\cite{zhong2017re}, which further boosts its mAP to 85.54\%.

\vspace{.1cm}
\noindent\textbf{PRID and iLIDS-VID:}
The comparisons on {PRID} and {iLIDS-VID} datasets are summarized in Table~\ref{table:compare2set}. It shows that, our method presents competitive performance on rank1 accuracy. M3D~\cite{li2019multi} also gets competitive performance on both datasets. The reason may be because the M3D jointly learns the multi-scale temporal cues form video sequences, and introduces a two-stream architecture to learn the spatial and temporal representations respectively. With a single feature extraction stream design, our method still outperforms M3D on both datasets. Table~\ref{table:compare2set} also compares with several temporal feature learning methods, \eg, RFA-Net~\cite{yan2016person}, SeeForest~\cite{zhou2017see}, T-CN~\cite{wu2018temporal}, CSA~\cite{chen2018video} and STMP~\cite{liu2019spatial}. Our method outperforms those works by large margins in rank1 accuracy.

\begin{table}
\footnotesize
\caption{Comparison with recent works on {PRID} and {iLIDS-VID}.}
\setlength{\tabcolsep}{9pt}
\label{table:compare2set}
\begin{tabular}{l|c|c|c|c}
\hline
Dataset&\multicolumn{2}{c|}{{PRID}}&\multicolumn{2}{c}{{iLIDS-VID}}\\
\hline
Method                              &rank1&rank5  &rank1&rank5\\
\hline
BoW+XQDA~\cite{zheng2016mars}       &31.80&58.50  &14.00&32.20\\
IDE+XQDA~\cite{zheng2016mars}       &77.30&93.50  &53.00&81.40\\
DFCP~\cite{li2017video}             &51.60&83.10  &34.30&63.30\\
AMOC~\cite{liu2017video}            &83.70&98.30  &68.70&94.30\\
QAN~\cite{liu2017quality}           &90.30&98.20  &68.00&86.80\\
DRSA~\cite{li2018diversity}         &93.20&-      &80.20&-\\
\hline
RCN~\cite{mclaughlin2016recurrent}  &70.00&90.00  &58.00&84.00\\
DRCN~\cite{wu2016deep}              &69.00&88.40  &46.10&76.80\\
RFA-Net~\cite{yan2016person}        &58.20&85.80  &49.30&76.80\\
SeeForest~\cite{zhou2017see}        &79.40&94.40  &55.20&86.50\\
T-CN~\cite{wu2018temporal}          &81.10&85.00  &60.60&83.80\\
CSA~\cite{chen2018video}            &93.00&99.30  &85.40&96.70\\
STMP~\cite{liu2019spatial}          &92.70&98.80  &84.30&96.80\\
M3D~\cite{li2019multi}              &94.40&100.00 &74.00&94.33\\
\hline
GLTR                                &{\bf95.50}&{\bf100.00}  &{\bf86.00}&\textbf{98.00}\\
\hline
\end{tabular}
\vspace{-2mm}
\end{table}

\vspace{.1cm}
\noindent\textbf{DukeMTMC-VideoReID:}
Comparisons on this dataset are shown in Table~\ref{table:compareduke}. Because DukeMTMC-VideoReID is a recently proposed video ReID dataset, a limited number of works have reported performance on it. We compare with ETAP-Net~\cite{wu2018exploit} and STA~\cite{fu2019sta} in this section. The reported performance of ETAP-Net~\cite{wu2018exploit} in Table~\ref{table:compareduke} is achieved with a supervised baseline.
As shown in Table~\ref{table:compareduke}, GLTR achieves 93.74\% mAP and 96.29\% rank1 accuracy, outperforming ETAP-Net~\cite{wu2018exploit} by large margins. The STA~\cite{fu2019sta} also achieves competitive performance on this dataset. GLTR still outperforms STA~\cite{fu2019sta} on rank1, rank5, and rank20 accuracy, respectively. Note that, STA~\cite{fu2019sta} utilizes extra body part cues and triplet loss.

\vspace{.1cm}
\noindent\textbf{Summary:}
The above comparisons on five datasets could indicate the advantage of GLTR in video representation learning for person ReID, \emph{i.e.}, it achieves competitive accuracy with simple and concise model design. We also observe that, the ReID accuracy on LS-VID is substantially lower than the ones on the other datasets. For example, the best rank1 accuracy on LS-VID is 63.07\%, substantially lower than the 87.02\% on MARS. This shows that, even though LS-VID collects longer sequences to provide more abundant spatial and visual cues, it still presents a more challenging person ReID task.

We show some person ReID results achieved by GLTR and ResNet50 baseline on LS-VID, MARS~\cite{zheng2016mars} and DukeMTMCVideoReID~\cite{wu2018exploit,ristani2016performance}  in Fig.~\ref{fig:result}. For each query, we show the top5 returned video sequences by those two methods. It can be observed that, the proposed GLTR is substantially more discriminative for identifying persons with similar appearance.

\begin{table}
\footnotesize
\caption{Comparison on {DukeMTMC-VideoReID}.}
\vspace{2mm}
\label{table:compareduke}
\setlength{\tabcolsep}{10pt}
\begin{center}
\begin{tabular}{l|c|c|c|c}
\hline
Method&mAP&rank1&rank5&rank20 \\
\hline
ETAP-Net~\cite{wu2018exploit} &78.34&83.62&94.59&97.58\\
STA~\cite{fu2019sta}          &\textbf{94.90}&96.20&{99.30}&99.60\\
\hline
GLTR             &93.74&\textbf{96.29}&\textbf{99.30}&\textbf{99.71}\\
\hline
\end{tabular}
\end{center}
\vspace{-4mm}
\end{table}

\section{Conclusion}
This paper proposes the Global Local Temporal Representation (GLTR) for video person ReID. Our proposed network consists of the DTP convolution and TSA model, respectively. The DTP consists of parallel dilated temporal convolutions to model the short-term temporal cues among adjacent frames. TSA exploits the relation among inconsecutive frames to capture global temporal cues. Experimental results on five benchmark datasets demonstrate the superiority of the proposed GLTR over current state-of-the-art methods.

~\\
\begin{small}
\textbf{Acknowledgments}
This work is supported in part by Peng Cheng Laboratory, in part by Beijing Natural Science Foundation under Grant No. JQ18012, in part by Natural Science Foundation of China under Grant No. 61620106009, 61572050, 91538111.
\end{small}

{\small
\bibliographystyle{ieee_fullname}
\bibliography{egbib}
}

\end{document}